\documentclass[lettersize,journal]{IEEEtran}
\usepackage{amsmath,amsfonts}
\usepackage{algorithmic}
\usepackage{algorithm}
\usepackage{array}
\usepackage[caption=false,font=normalsize,labelfont=sf,textfont=sf]{subfig}
\usepackage{textcomp}
\usepackage{stfloats}
\usepackage{url}
\usepackage{verbatim}
\usepackage{graphicx}
\usepackage{cite}
\usepackage{algorithm}
\usepackage{algorithmic}
\usepackage{multirow}
\usepackage{makecell}
\usepackage{ragged2e}
\usepackage{amssymb}
\usepackage{amsmath}
\usepackage{booktabs}
\usepackage{threeparttable}
\usepackage{graphicx}
\usepackage{newfloat}
\usepackage{listings}
\usepackage{color}
\begin{document}

\title{GroupVideo: Multi-Identity Customized Text-to-Video Generation}

\author{Xinyang Song\IEEEmembership{,~Graduate Student Membder,~IEEE}, Libin Wang, Jianxin Sun, Qi Li\IEEEmembership{,~Member,~IEEE}, \\Dandan Zheng, JingDong Chen, Zhenan Sun\IEEEmembership{,~Member,~IEEE}

\thanks{This work was supported in part by the National Key Research and Development Program of China under Grant 2024YFC3210803, in part by the National Natural Science Foundation of China (Grant Nos. U23B2054, 62276263), and the Youth Innovation Promotion Association CAS (Grant No. Y2023143). 
\textit{(Corresponding author: Qi Li.)}}
\thanks{Xinyang Song, Qi Li and Zhenan Sun are with the School of Artificial Intelligence, University of Chinese Academy of Sciences, Beijing 100049, China, and also with the New Laboratory of Pattern Recognition and the State Key Laboratory of Multimodal
Artificial Intelligence Systems, Institute of Automation, Chinese Academy of Sciences, Beijing 100190, China (email: xinyang.song@cripac.ia.ac.cn, qli@nlpr.ia.ac.cn, znsun@nlpr.ia.ac.cn)}
\thanks{Libin Wang, Jianxin Sun, Dandan Zheng and Jingdong Chen are with Ant Group, Beijing 100081, China (email: libin.wlb@antgroup.com, sun jianxin.sjx@antgroup.com, yuandan.zdd@antgroup.com, jingdongchen.cjd@ antgroup.com)}
}



\maketitle

\begin{abstract}
Current identity customized video generation methodologies are predominantly limited to single-identity scenarios, as the lack of explicit identity separation mechanisms often leads to identity confusion in multi-identity settings. 
Existing multi-identity approaches, which directly extend single-identity frameworks by concatenating face images as input conditions, frequently result in unnatural facial expressions and motions, manifesting as the ``copy-paste" phenomenon. 
To overcome these limitations, we introduce GroupVideo, a novel framework that leverages multiple individual photographs to generate identity-customized video.
Built upon Video Diffusion Transformers, GroupVideo incorporates multimodal identity alignment: visual alignment jointly encodes multiple face images to provide robust identity references, while semantic alignment introduces a semantic perceiver to enhance the naturalness of motions.
An ID localization module with spatial guidance is introduced to address identity blending and enhance identity fidelity, along with bounding box constraints and mask regularization loss, to focus on facial regions and improve training efficiency.
In response to the shortage of multi-ID video datasets, we have curated a comprehensive high-quality dataset of 20,000 videos, thereby establishing a crucial resource to advance future research in multi-ID video generation.
Extensive experiments demonstrate that GroupVideo outperforms existing methods in generating multi-character videos with consistent identities and natural motions. 
\end{abstract}

\begin{IEEEkeywords}
Video Customization, Identity Preservation, Text-to-Video Generation
\end{IEEEkeywords}

\section{Introduction}
The rapid advancement of video diffusion models, particularly Diffusion Transformers \textcolor{black}{(DiTs)}, has fundamentally transformed content creation capabilities from personalized video synthesis~\cite{ruiz2023dreambooth, gal2022image} to identity-aware generation~\cite{yuan2024identity, he2024id}. Within this landscape, multi-identity customized video generation emerges as a critical challenge: synthesizing temporally coherent videos that faithfully preserve multiple identities while executing text-driven actions.
\textcolor{black}{This task requires the model to simultaneously maintain identity fidelity, motion naturalness, and spatial-temporal consistency across multiple characters.}

\textcolor{black}{Recent progress in ID-consistent video generation has led to a series of efforts for personalized content creation. However, many existing methods rely on inflated U-Net fine-tuning or adapter-based optimization, which limits their scalability and flexibility.}
Recent works~\cite{yuan2024identity, wei2025echovideo, zhang2025magic, fei2025ingredients} have pivoted towards \textcolor{black}{offline-training} ID-preserving video generation methods~\cite{zhang2025magic, yuan2024identity, wei2025echovideo} based on Diffusion Transformers~\cite{peebles2023scalable}. 
\textcolor{black}{Among them, ConsisID~\cite{yuan2024identity} is a representative framework for high-fidelity identity-preserving generation. Nevertheless, it is mainly designed for \textbf{single-ID customization}.}

\begin{figure}[t]
  \centering
   \includegraphics[width=1.0\linewidth]{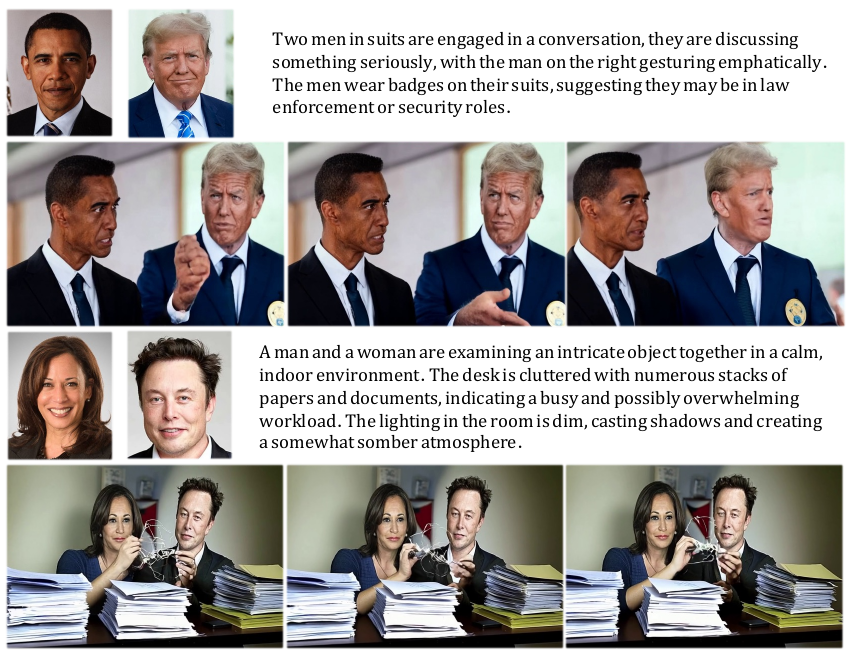}
   \caption{Existing multi-character generation methods, \textcolor{black}{such as Ingradients~\cite{fei2025ingredients},} suffer from (1) \textbf{weak identity preservation}, (2) \textbf{stiff, static motions (copy-paste)}, and (3) semantic conflicts causing \textbf{unnatural distortions}.}
   \label{fig:2}
\end{figure}

\textcolor{black}{A straightforward extension to multi-ID generation is to concatenate multiple identity conditions. For example, Ingredients~\cite{fei2025ingredients} extends ConsisID by concatenating two facial conditions for multi-ID customization. However, such a design still follows a simple condition-stacking paradigm and often suffers from a pronounced \textbf{``copy-paste'' effect}, leading to \textbf{rigid expressions, limited motion diversity, and identity interference} among different characters, as illustrated in Fig.~\ref{fig:2}. These limitations suggest that multi-ID customized video generation is not merely a direct extension of single-ID generation. Instead, it requires a more dedicated mechanism to jointly model multiple identities, their semantic roles, and their spatial relationships in dynamic video scenes.}

To overcome these challenges, we introduce GroupVideo, a novel unified framework designed to bridge the gap between multi-ID customization and high-quality video synthesis.
Specifically, we propose a \textbf{multimodal identity alignment mechanism}, which effectively captures multi-scale identity features from facial images and seamlessly integrates them into both the visual latent space and textual semantic space.
\textcolor{black}{In the visual branch, instead of simply concatenating reference images, we extend the visual channels to achieve identity-aware visual alignment, which helps avoid static and rigid visual artifacts. In the semantic branch, ID embeddings are fused with textual embeddings through a semantic perceiver, enabling the model to establish semantic correspondence between identity cues and action descriptions, thereby facilitating more coherent human motions.}
Notably, this design is inherently scalable, supporting an arbitrary number of identities. 

\IEEEpubidadjcol

\textcolor{black}{Another major challenge in multi-ID generation is the \textbf{identity ambiguity} caused by simultaneously injecting multiple identity features. To alleviate this issue, we further introduce an \textbf{ID localization module}, which dynamically models the correlations among multiple identity embeddings based on attention maps and provides implicit spatial guidance during generation. Compared with simple global identity conditioning, this module is better suited for multi-character scenes, where accurate identity-position association is crucial for preserving character consistency across frames.}

In order to achieve efficient optimization, a \textbf{progressive two-stage training strategy} is adopted to optimize the aforementioned modules separately.
 To further enhance identity consistency, we incorporate \textbf{a bounding box constraint and a mask regularization loss} in each training stage, thereby reducing the effect of non-facial regions during optimization.
 Recognizing the scarcity of high-quality multi-person video datasets, we have collected and curated \textbf{a high-resolution multi-person video dataset} comprising 20,000 samples.
 Each video has been annotated with detailed captions and carefully filtered based on the number of individuals present and the proportion of faces visible.
 We believe the dataset will facilitate future research in multi-ID customized video generation.
\IEEEpubidadjcol

\begin{figure*}[t]
    \centering
    \includegraphics[width=1.0\linewidth]{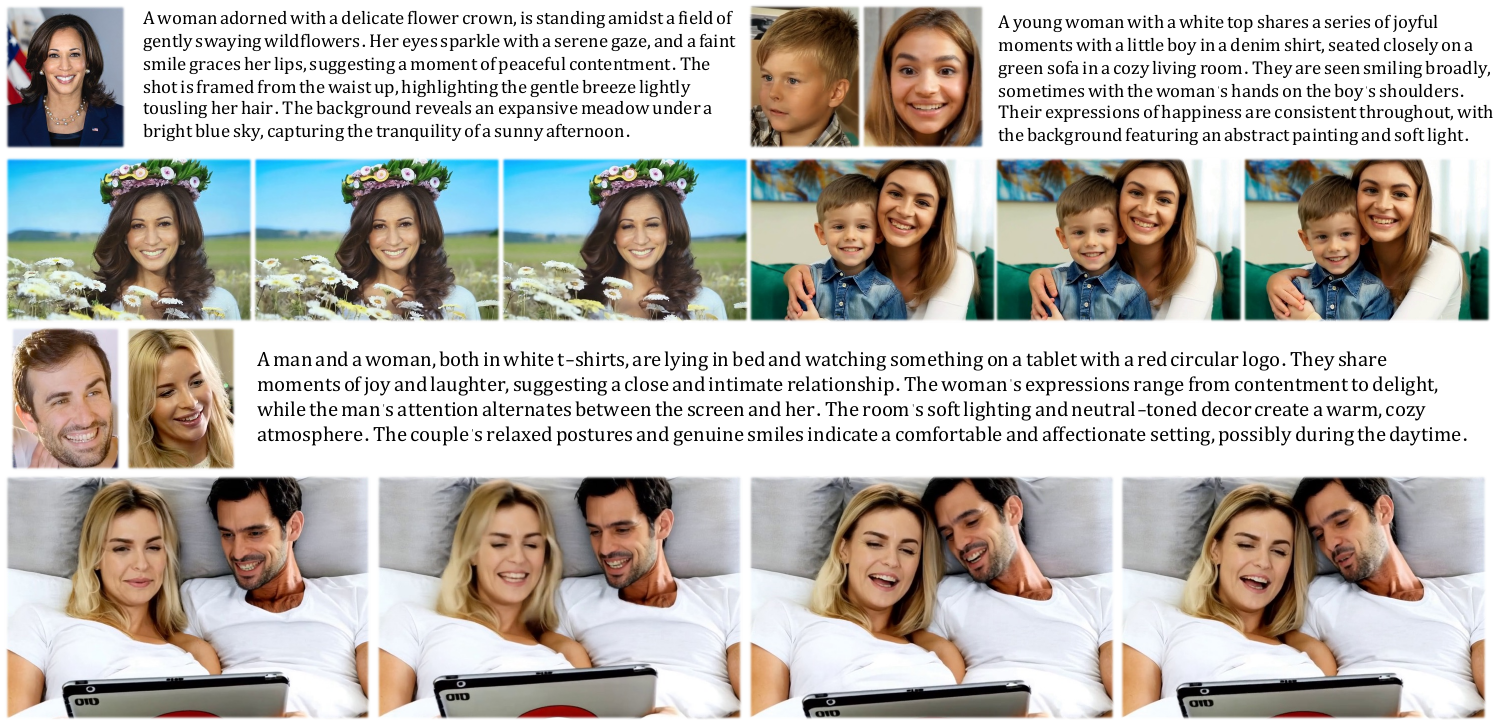}
    \caption{Sample results of the proposed GroupVideo. Given facial reference images, GroupVideo generates \textbf{identity-consistent} videos featuring either single or multiple characters, characterized by \textbf{natural motions} and \textbf{precise alignment with textual descriptions}.}
    \label{fig:1}
\end{figure*}

Our contributions can be summarized as follows: 
\begin{itemize}
\item[$\bullet$] 
We propose GroupVideo, \textcolor{black}{an offline-training} multi-ID customized video generation framework, achieving robust identity consistency across multiple individuals while generating natural and realistic human motions.
\item[$\bullet$] 
\textcolor{black}{We introduce a novel multimodal identity alignment mechanism that jointly integrates identity information into the visual latent space and textual semantic space, together with an ID localization module that mitigates identity ambiguity under multi-ID conditioning through implicit spatial guidance.}
\item[$\bullet$]
\textcolor{black}{We develop a progressive two-stage optimization strategy with bounding box loss and mask regularization to improve training stability and identity consistency, and we further contribute a high-quality multi-person video dataset of 20,000 videos to support future research in multi-character video generation.} 
\end{itemize}

\section{Related Works}
\label{sec:related}

\subsection{Video Diffusion Models}
Diffusion models have emerged as the dominant paradigm in image and video generation, surpassing traditional approaches based on GANs~\cite{goodfellow2020generative, karras2019style, sounding, controllable} and autoregressive models~\cite{esser2021taming, yu2022scaling}, owing to their superior diversity and controllability.
The evolution of text-to-image diffusion models~\cite{rombach2022high, saharia2022photorealistic, podell2023sdxl, song2026unialignment, song20253sgen, song2026fine}, initially built on inflated U-Net architectures, aid the foundation for early video generation models~\cite{benchmark, ta2v}.
Recent advancements have transitioned from U-Net backbones to DiTs, incorporating 3D spatio-temporal attention mechanisms.
This paradigm enhances the model's capacity to capture the complex video dynamics and ensures seamless frame transitions.
The emergence of SORA~\cite{videoworldsimulators2024} and Latte~\cite{ma2024latte} has demonstrated the significant potential of DiT in improving video quality, sparking a wave of DiT-based video generation research~\cite{lin2024open, zheng2024open, yang2024cogvideox, kong2024hunyuanvideo, lin2025toklip, li2026iv, yang2026conceptguided}.
Building upon these advancements, our work investigates the controllability of video diffusion transformers, with a particular emphasis on identity preservation in video generation.


\subsection{Identity preserving Generation}
\textbf{Single-ID Customization.} 
Identity-preserving generation focus on synthesizing images or videos that maintain consistent character identities based on a given reference image.
Early approaches to identity-specific synthesis~\cite{ruiz2023dreambooth, ruiz2024hyperdreambooth, gal2022image, hu2021lora, sgdm, multiview} relied on fine-tuning the entire network or subsets of parameters during inference using one or more reference images, incurring substantial computational overhead. Recent \textcolor{black}{offline-training} methods\cite{wang2024instantid, ye2023ip, guo2024pulid, li2024blip} have addressed this limitation by leveraging pre-trained encoders to extract and align identity features without parameter updates. Another line of research~\cite{li2024photomaker, chen2024dreamidentity, gal2023encoder} concatenates identity features with text embeddings for conditional generation.
Unlike portrait video generation~\cite{xie2024x, hu2024animate, he2025posgen} and image animation~\cite{guo2023animatediff, zhang2023i2vgen, xing2024dynamicrafter}, which primarily focus on animating a given character's image, ID-customized video generation~\cite{ma2024magic, he2024id, chefer2024still, wei2024dreamvideo, jiang2024videobooth, yuan2024identity, wei2025echovideo, zhang2025magic} constructs novel scenes and actions while maintaining strict identity consistency. U-Net-based frameworks have adopted divergent strategies: MagicMe~\cite{ma2024magic} employs fine-tuning with reference image, while ID-Animator~\cite{he2024id} introduces a face adapter with decoupled cross-attention layers for parameter-free operation. However, these approaches often struggle with facial similarity and overall video quality. In contrast, methods~\cite{zhang2025magic, wei2025echovideo, yuan2024identity} based on the DiT architecture achieve superior scalability and video quality.

\noindent\textbf{Multi-ID Customization.} 
In contrast to single-ID customization, multi-ID customization introduces the significant challenge of identity blending, where multiple identities must coexist naturally within the same scene.
Conventional solutions~\cite{kumari2023multi, liu2023cones, videodreamer, animediff} address this by encoding subject information into text embeddings, relying on semantic descriptions to differentiate identities.
While intuitive, these methods impose rigid constraints on prompt formats and often sacrifice identity fidelity or text consistency due to the entanglement between visual and textual embeddings. 
Another line of research~\cite{kim2024instantfamily, gu2024mix, xiao2024fastcomposer, stableidentity} employs predefined spatial masks to provide explicit positional guidance.
 Although effective in reducing identity overlap, these methods require precise layout specifications, limiting their applicability in dynamic scenarios. Furthermore, the reliance on fixed masks often compromises the diversity and naturalness of generated content. 
Recent adavancements~\cite{wang2024moa, he2024uniportrait} further decouple identity embeddings using ID-Router modules, specifying ID locations in a self-supervised manner. 
Ingredients~\cite{fei2025ingredients} represents the first attempt at multi-ID video customization using DiT architectures. However, its naive conditioning strategy results in severe ``copy-paste" artifacts, where characters appear unnaturally superimposed with minimal interaction or motion coherence.
In this work, we propose a multimodal identity alignment mechanism and an ID localization module, effectively addressing the aforementioned challenges.

\begin{figure*}[t]
  \centering
  \includegraphics[width=1.0\textwidth]{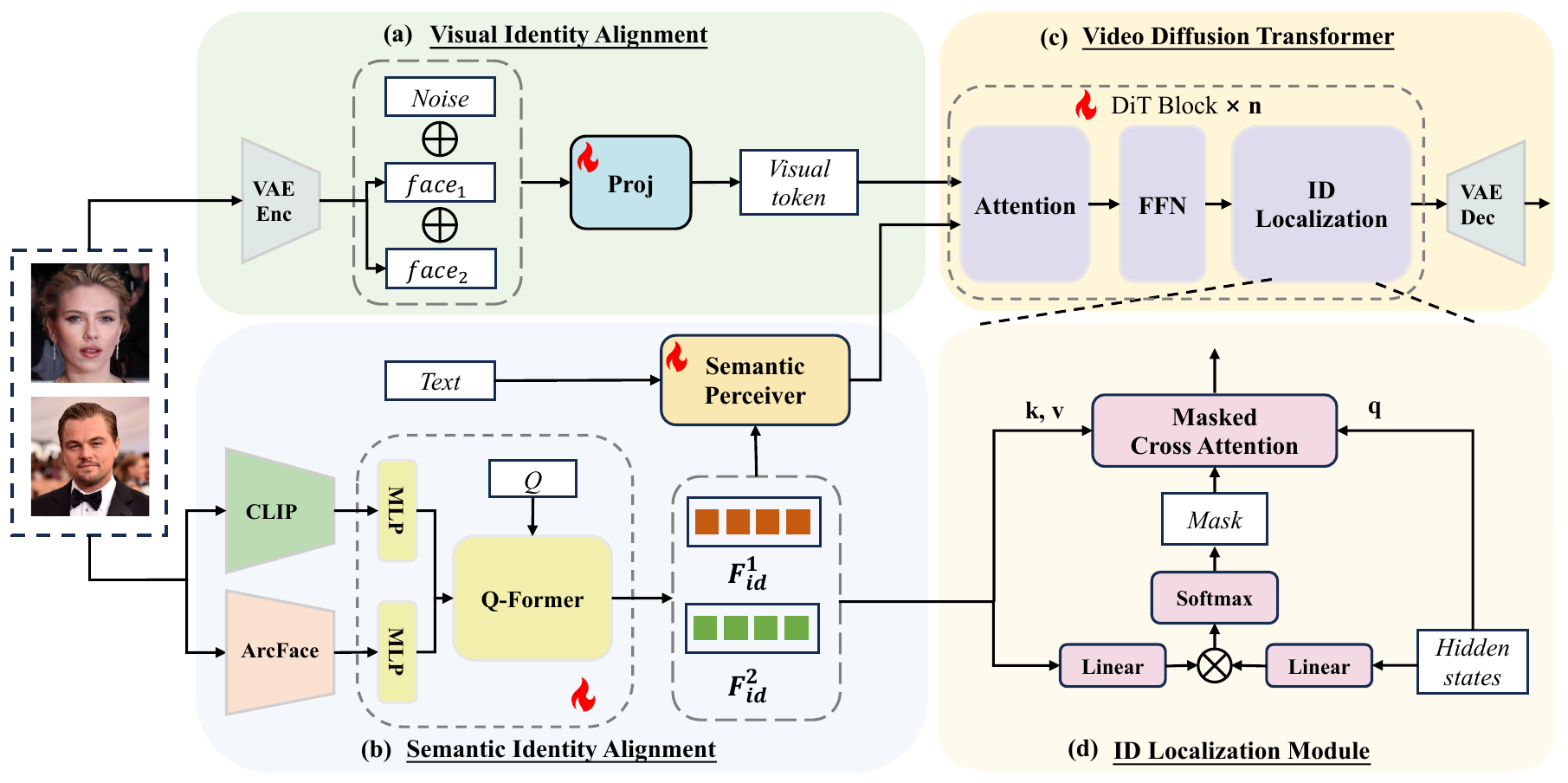}
   \caption{\textbf{Overview of GroupVideo.} GroupVideo introduces multimodal identity alignment based on (c) Video Diffusion Transformer. The (a) visual identity alignment and (b) semantic identity alignment fuse multiple identity features with visual latents and semantic embeddings respectively. This is followed by an (d) ID localization module to route different identities, which is illustrated on the lower right. By optimizing the trainable parameters marked in the diagram, GroupVideo effectively integrates multiple facial features for controllable video synthesis while ensuring identity fidelity across the generated sequence. }
   \label{fig:3}
\end{figure*}

\section{Methodology}
This section introduces GroupVideo, an innovative approach to multi-ID character video generation based on video diffusion transformer. 
An overview of GroupVideo is illustrated in Fig.~\ref{fig:3}. 
We begin by providing a brief review of diffusion models. 
The multimodal identity alignment mechanism extracts latent representations from character images as visual embeddings and establishes semantic associations between identity and text embeddings. 
This not only provides robust multi-character references but also mitigates the occurrence of rigid, static scene.
The identity embeddings are further processed by the ID localization module, which provides position-wise activation strengths for the attention mask, effectively preventing identity blending.
Finally, we illustrate the progressive training strategy and the two designed loss functions.

\subsection{Preliminaries}
\label{sec:3.1}
Diffusion models operate by iteratively reversing a noise perturbation process, transforming random noise into a generated clean sample through a series of denoising steps. 
The model predicts the latent distribution $z_t$ at the current time step based on $z_{t+1}$ using a Gaussian transition:
\begin{equation}
    q_\theta\left(z_{t} \mid z_{t+1}\right)=\mathcal{N}\left(z_{t} ; \mu_\theta\left(z_{t+1}, t\right), \sigma_t\right),
\end{equation}
where $\theta$ represents the learnable parameters, $\mu_\theta$ is the predicted mean, and $\sigma_t$ is the variance schedule.
Latent Diffusion Models (LDMs)~\cite{rombach2022high} further improve efficiency by shifting the denoising process from pixel space to latent space, significantly reducing computational overhead.
The noise prediction process is optimized by minimizing a mean squared error (MSE) loss: 
\begin{equation}
\label{equ:2}
    \mathcal{L}_{\text{MSE}}=\mathbb{E}_{x_0, t, y, f, \epsilon}\left[\left\|\epsilon-\epsilon_\theta\left(x_0, t, \varphi(y), \tau(f)\right)\right\|_2^2\right],
\end{equation}
where $\epsilon$ denotes the predicted noise, $y$ and $f$ represent text and image respectively when both conditions are supplied.
In our DiT-based video generation model, the text condition $c_{text}$ is obtained by a T5 encoder~\cite{raffel2020exploring}, while the image condition $c_{img}$ is derived by processing the face image $f$ through a feature extractor $\tau$.

\subsection{MultiModal Identity Alignment}
\label{sec:3.2}
In this work, we propose a decoupled approach for multimodal identity alignment, where identity features extracted from facial regions are independently fused with both visual latents and textual embeddings. This ensures that the model retains precise visual information while preserving facial semantic details. Given a set of input images $I_{in}$ containing facial regions, we first employ a face extractor to isolate the facial images. The visual identity alignment branch (the upper left part of Fig.~\ref{fig:3}) processes these facial images through a VAE to obtain visual tokens, which provide a compact yet effective representation of identities. Unlike~\cite{fei2025ingredients} that concatenates multiple facial images in pixel space to form composite images, our method concatenates multiple visual tokens with noise latent along the channel dimension. The concatenated tokens are then projected through a patch embedding layer and serves as the input to the DiT block.

By treating multiple facial images as independent conditional inputs, our approach provides balanced control signals without constraining the relative spatial relationships of individuals in the generated videos. This design eliminates the generation of rigid, static scenes and instead fosters richer diversity in the output. The decoupled nature of our method ensures that each facial image contributes equally to the generation process, enhancing the model's ability to produce dynamic and visually coherent results.

The semantic identity alignment branch (the lower left part of Fig.~\ref{fig:3}) builds a fine-grained semantic bridge between identity embeddings and text embeddings. Unlike existing methods that employ separate cross-attention mechanisms~\cite{he2024uniportrait, wang2024instantid} or non-interactive approaches~\cite{yuan2024identity},\textcolor{black}{\cite{cheng2026prompt, cheng2026isolating, cheng2025semantic}}, to independently inject these two conditions, our approach addresses the challenge of establishing meaningful associations between identity and textual descriptions. This is crucial to avoid inconsistent results when role attributes in the prompt change. Similar to~\cite{guo2024pulid, yuan2024identity}, we leverage a face recognition backbone~\cite{deng2019arcface} to extract intrinsic identity attributes while utilizing the CLIP encoder to capture rich semantic information. These two streams are integrated through a multi-modal vision encoder Q-Former~\cite{li2023blip}, which transforms learnable embeddings into intrinsic identity features $F_{id}^n=[F_{id}^1, ..., F_{id}^N]$, $n=1,...,N$, where $N$ represents the character number. 

Once the identity features are obtained, we employ a semantic perceiver to interact with the textual embeddings, enhancing the fusion of non-identity attributes and identity features via cross-attention:
\begin{equation}
\scalebox{1.0}{$
F_{t} = F_{t} + \sum_i^N \mathrm{Attention}(Q^{t}, K^{f}_i, V^{f}_i),
$}
\end{equation}
where $Q^t=F_{t}W^q$, $K^f_i=F_{id}^iW^k$ and $V^f_i=F_{id}^iW^v$. $F_{t}$ denotes the text embedding, $W_q$, $W_k$ and $W_v$ are learnable parameters. The aligned textual embeddings, combined with the visual embeddings, provide a joint guidance direction for identity-consistent generation. The semantic identity alignment maintains semantic coherence and identity fidelity, even with significant changes in textual prompts.

\subsection{ID Localization Module}
\label{sec:3.3}
Through multimodal identity alignment, multiple independent entity-aware embeddings collectively provide identity information. However, directly inputting ID embeddings tends to cause identity confusion among characters. To prevent identity blending, we introduce an ID Localization Module, which offers spatial guidance for multiple identity features, as illustrated in the lower right part of Fig.~\ref{fig:3}. This module learns a localization map to predict the effective regions for each identity embedding, thereby assigning each identity to its corresponding spatiotemporal location within the latent space. 

Specifically, we first compute a localization mask $\mathrm{M}$ for each ID using \textcolor{black}{hidden states $H\in \mathbb{R}^{1 \times l_H \times c_H}$ (the intermediate feature of the DiT block) and multiple identity embeddings $F_{id}^n \in \mathbb{R}^{1 \times l_{id} \times c_{id}}$, $n=1, ..., N$:}
\begin{equation}
\scalebox{1.0}{$
\label{equ:4}
    \mathrm{M}_{n} = \mathrm{Softmax}(\phi(H)*\theta(F_{id}^n)),
$}
\end{equation}
\textcolor{black}{
where $\phi$ and $\theta$ are two linear projection layers, used for feature-dimension alignment and sequence-length compression, respectively. 
}
$\mathbf{*}$ denotes matrix multiplication operator, and the $\mathrm{Softmax}$ function is applied to convert the computed weights into an N-dimensional probability distribution, indicating the likelihood of spatial-temporal positions for multiple ID features.

\textcolor{black}{
The localization masks $\mathrm{M}_n \in \mathbb{R}^{1 \times l_H}$ are concatenated together as a condition to route identity embeddings through masked cross-attention modules:
}
\begin{equation}
\scalebox{1.0}{$
    \psi_i(H, F_{id}) = [\operatorname{M}_n]_{n=1}^N * \operatorname{Attention}_i(H, F_{id}).
$}
\end{equation}
The predicted mask is used to provide spatial guidance weights for the subsequent cross-attention, enhancing identity awareness in the cross-attention mechanism while being more computationally efficient than attention maps.
\textcolor{black}{
The output $\psi_i(H, F_{id}) \in \mathbb{R}^{N \times l_H \times c_H}$ denotes a unified representation in which each fused attention result is restricted to the spatiotemporal region specified by its corresponding mask.
}

This mechanism ensures that each identity feature is spatially localized, preventing overlap and preserving the distinctiveness of each identity. 
The ID localization module is integrated into multiple DiT blocks, enabling dynamic adjustment of facial feature localization during the generation process. This integration allows the model to adaptively refine the spatial distribution of identity features at different stages of the generative pipeline, ensuring precise and coherent placement of multiple identities in the final output.

\begin{figure}[t]
  \centering
  \includegraphics[width=1.0\linewidth]{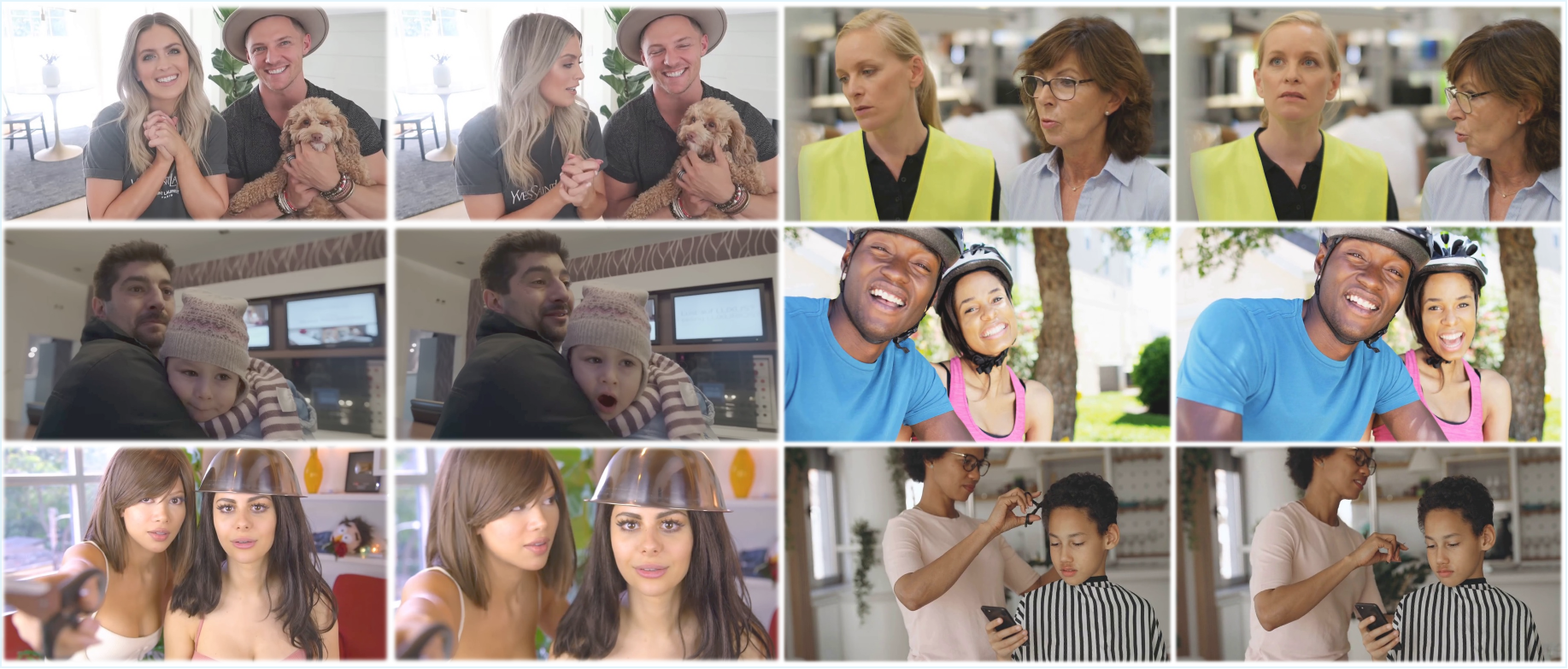}
  \caption{\textcolor{black}{Examples from the multi-character video dataset. The dataset comprises a total of 20,000 videos, encompassing a diverse range of multi-angle, high-resolution portraits.}}
  \label{fig:4}
\end{figure}

\subsection{Progressive Training Strategy}
\label{sec:3.4}
We adopt a progressive two-stage training pipeline to ensure robust and efficient learning. In the first stage, we focus on multimodal identity alignment training. \textcolor{black}{During this phase, the semantic perceiver, ID embedding extraction module, and the entire parameters of the pre-trained video generation backbone are jointly optimized.}
To mitigate interference from redundant background information, the optimization objective in the first stage incorporates \textcolor{black}{a bounding box constraint $\mathcal{L}_{box} = \mathbf{B} \otimes \mathcal{L}_{mse}$}, in addition to the diffusion loss in~\eqref{equ:2}:
\textcolor{black}{
\begin{equation}
\scalebox{1.0}{$
\mathcal{L}_{s1}= 
\begin{cases}
\mathcal{L}_{mse}, & \text { if } p>\alpha \\
\mathcal{L}_{box}, & \text { if } p \leq \alpha
\end{cases}
$}
\end{equation}
}
\textcolor{black}{where $\mathbf{B}$ represents the compressed bounding boxes of the facial regions in the video, $p\sim \mathcal{U}(0,1)$, and $\alpha$ controls the probability of applying the bounding box constraint. This encourages the model to concentrate more on the facial regions.}

In the second stage, we fine-tune only the ID localization module while keeping all other parameters fixed. This operation ensures finer differentiation among multiple identities. To further facilitate the module's ability to distinguish and localize different subject instances, we introduce a mask regularization loss:
\textcolor{black}{
\begin{equation}
\scalebox{1.0}{$
    \mathcal{L}_{\text {mask}}=\frac{1}{N} \sum_{i=1}^N\left\|\mathbf{M}_i-\mathbf{M}_i^{g t}\right\|_2^2,
$}
\end{equation}
}
where $\mathbf{M}_i^{gt}$ denotes the downsampled ground truth face mask sequence for the i-th person in the target video, and \textcolor{black}{$\mathbf{M}_i$} is the predicted mask in \eqref{equ:4}. The mask regularization loss explicitly enhances the mask prediction accuracy, meanwhile avoiding large iteration steps for learning mask prediction, which accelerates convergence in early stages. The overall loss function of the second stage is defined as: 
\begin{equation}
    \mathcal{L}_{s2} = \mathcal{L}_{mse} + \lambda \mathcal{L}_{mask}
\end{equation}
This two-stage training strategy ensures that the model first learns robust identity representations and then refines its ability to spatially localize multiple identities, leading to more accurate and coherent video generation.

\begin{table}
\setlength{\tabcolsep}{10pt}
    \centering
    \caption{\textbf{Comparison of ID-consistent video datasets.} GroupVideo offers multi-person composition with full-body coverage, high-resolution inputs, and open accessibility.}
    \label{table:1}
{\normalsize
    \begin{tabular}{@{}lccc@{}}
    \toprule
    Datasets            & \makecell{Body Coverage}   & Resolution    & Ids  \\
    \midrule
    CelebV-Text & Face      & 512$\times$512 & 1    \\
    ID-Animator & Face      & 512$\times$512 & 1    \\
    EchoVideo   & Face      & 480$\times$720 & 1    \\
    ConsisID    & Face\&Body    & 480$\times$720 & 1 \\
    Ingredients & Face\&Body    & 480$\times$720  &2   \\
    \textbf{GroupVideo} & Face\&Body  & \textbf{1080$\times$1920} &\textbf{2$\sim$3}  \\
    \bottomrule
    \end{tabular}
}
\end{table}

\begin{figure*}[ht!]
  \centering
  \includegraphics[width=0.95\textwidth]{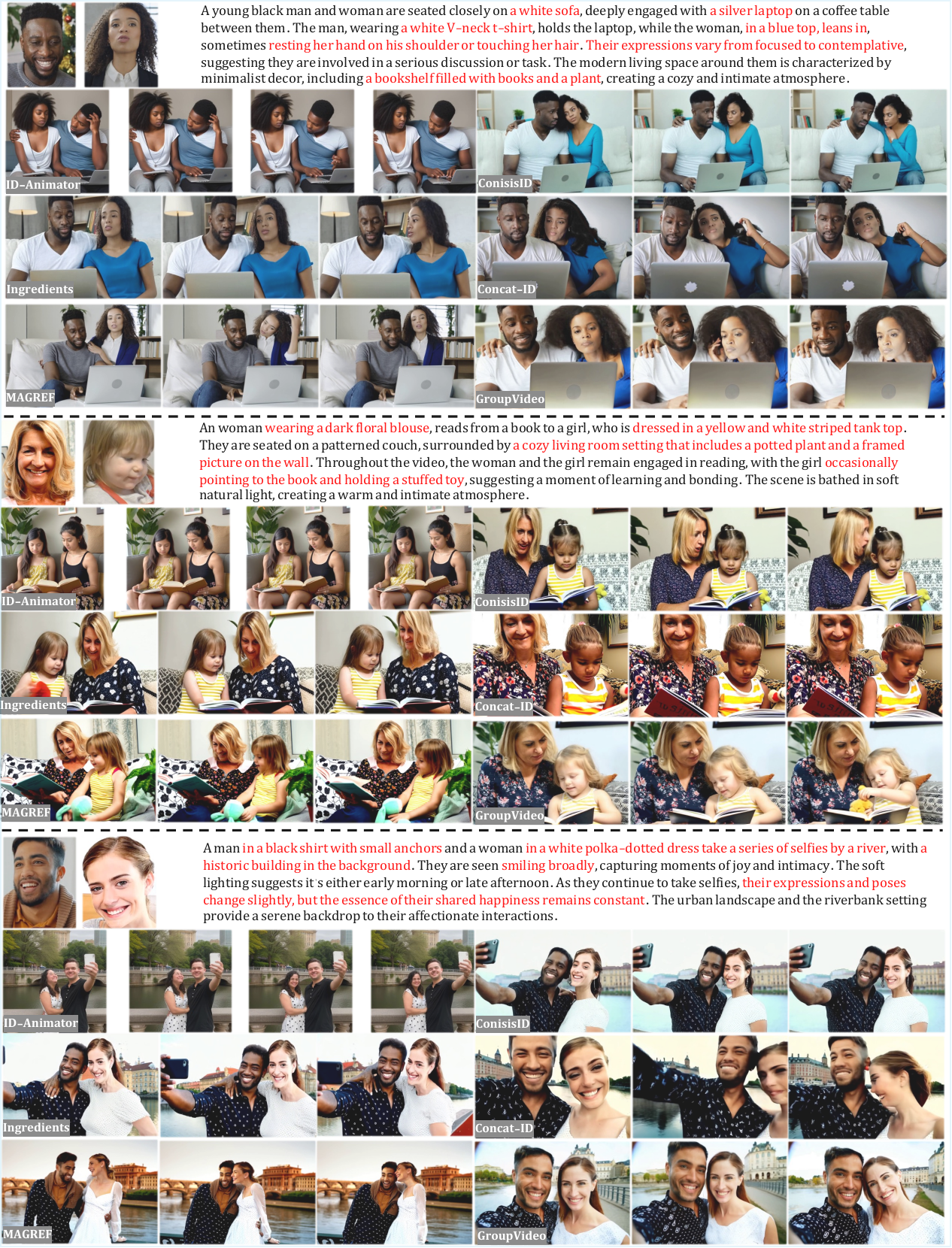}
  \caption{\textcolor{black}{\textbf{Qualitative comparison with state-of-the-art ID-preserved video generation methods.} The identity references are displayed on the top-left, the attributes in the textual instructions are highlighted in red.}}
  \label{fig:5}
\end{figure*}

\begin{table*}
\setlength{\tabcolsep}{4.5pt}
    \centering
    \caption{\textbf{Quantitative evaluation of ID customized video generation.} The best results are highlighted in bold.} 
    \label{table:2}
{\normalsize
\begin{tabular}{llccccc}
\toprule
\multirow{2}{*}{Method} &  \multirow{2}{*}{IDs}  & \multicolumn{3}{c}{Visual quality}    & \multicolumn{2}{c}{Conditional Consistency} \\   
\cmidrule(r){3-5}   \cmidrule(r){6-7}  & &FID~\cite{heusel2017gans}$\downarrow$  &Aesthetic Quality~\cite{huang2024vbench}$\uparrow$  &Dynamic Degree~\cite{huang2024vbench}$\uparrow$ & FaceSim~\cite{schroff2015facenet}$\uparrow$  & CLIPScore~\cite{hessel2021clipscore}$\uparrow$   \\  
\cmidrule(r){1-2}  \cmidrule(r){3-7}  
ID-Animator~\cite{he2024id} &\multirow{6}{*}{2 ID} & 218.327 & 0.513 & 0.311 & 0.102 & 25.728    \\ 
ConsisID~\cite{yuan2024identity} & & 162.547 & 0.605 & 0.787 & 0.634 & \underline{31.245} \\ 
Ingredients~\cite{fei2025ingredients} & & 165.541    & 0.587  & 0.757  & \textbf{0.644} & 30.220  \\ 
\textcolor{black}{Concat-ID~\cite{zhong2025concat}} & & \textcolor{black}{170.366} & \textcolor{black}{0.601} & \textcolor{black}{0.733} & \textcolor{black}{0.625} & \textcolor{black}{31.108} \\
\textcolor{black}{MAGREF~\cite{deng2025magref}}  & & \textcolor{black}{\underline{151.490}} & \textcolor{black}{\underline{0.609}} & \textcolor{black}{\textbf{0.828}} & \textcolor{black}{0.630} & \textcolor{black}{30.845}\\
\textbf{GroupVideo} & &\textbf{147.228} & \textbf{0.611} & \underline{0.821} & \underline{0.636} &  \textbf{32.035}  \\ \cmidrule(r){1-2} 
\cmidrule(r){3-7}
ConsisID~\cite{yuan2024identity} &\multirow{2}{*}{1 ID} & 156.336 & \textbf{0.602} & 0.762 & \textbf{0.653} & 31.788 \\ 
\textbf{GroupVideo} & & \textbf{150.314} & 0.596 & \textbf{0.830} & 0.648 & \textbf{32.114} \\ 
 \bottomrule
\end{tabular}
}
\end{table*}

\section{Experiments}

\subsection{Multi-Character Video Dataset}
Current datasets for ID-preserving video generation suffer from three primary limitations: (1) restricted body part coverage (typically facial regions only), (2) single-identity composition, and (3) low-resolution training data with artificial artifacts. As evidenced in Tab.~\ref{table:1}, existing solutions either limit their scope to individual faces~\cite{yu2022celebvtext,he2024id} or lack open accessibility~\cite{wei2025echovideo}. While recent efforts~\cite{yuan2024identity} attempt full-body generation, they remain constrained to single-person scenarios. Ingredients~\cite{fei2025ingredients} partially addresses multi-identity needs but suffers from severe resolution limitations ($480\times 720$) and text overlay contamination in its 1,000 video clips.

To overcome these constraints, we construct  a high-quality dataset containing 20,000 HD video clips ($1080\times 1920$) featuring 2-3 interacting identities. 
\textcolor{black}{
Fig.~\ref{fig:4} shows examples from the multi-character video dataset.
To thoroughly focus on ID preservation, we ensure that the dataset contains videos with clearly visible facial regions and covers diverse ages and genders.
}
Our collection pipeline employs rigorous multi-stage filtering: (1) We retrieve candidate videos from diverse online sources, ensuring native HD resolution. (2) Videos with subtitles or watermarks are excluded to avoid unnecessary noise. (3) Optical flow and aesthetic scores are calculated to filter out videos with low motion dynamics and poor visual quality. (4) We then perform joint vision-language verification utilizing Qwen-VL~\cite{bai2023qwenvlversatilevisionlanguagemodel} and face detection to select clips containing exactly 2-3 discernible characters. 
\textcolor{black}{(5) We use YOLO~\cite{redmon2016you} to extract face bounding boxes for each frame in a clip, and apply SAM-2~\cite{ravi2024sam} to the highest-confidence face crop to generate the corresponding face masks. These annotations are stored as ground truth for the bounding box constraint and the mask regularization loss. Based on the detected face bounding boxes, we compute the duration and the area proportion of facial regions within each segment to ensure sufficient on-screen presence and visibility of the characters. In addition, we discard samples where the face regions are too small or difficult to detect.}
(6) A video understanding model, CogVLM~\cite{wang2023cogvlm} is utilized to generate detailed captions for each video, describing the content, actions, and interactions of the characters. 
Benefiting from the aforementioned procedures, our dataset ensures clear character faces, uniquely combining full-body coverage, multi-identity support, and open accessibility at unprecedented scale.

\begin{figure*}[t]
  \centering
   \includegraphics[width=1.0\linewidth]{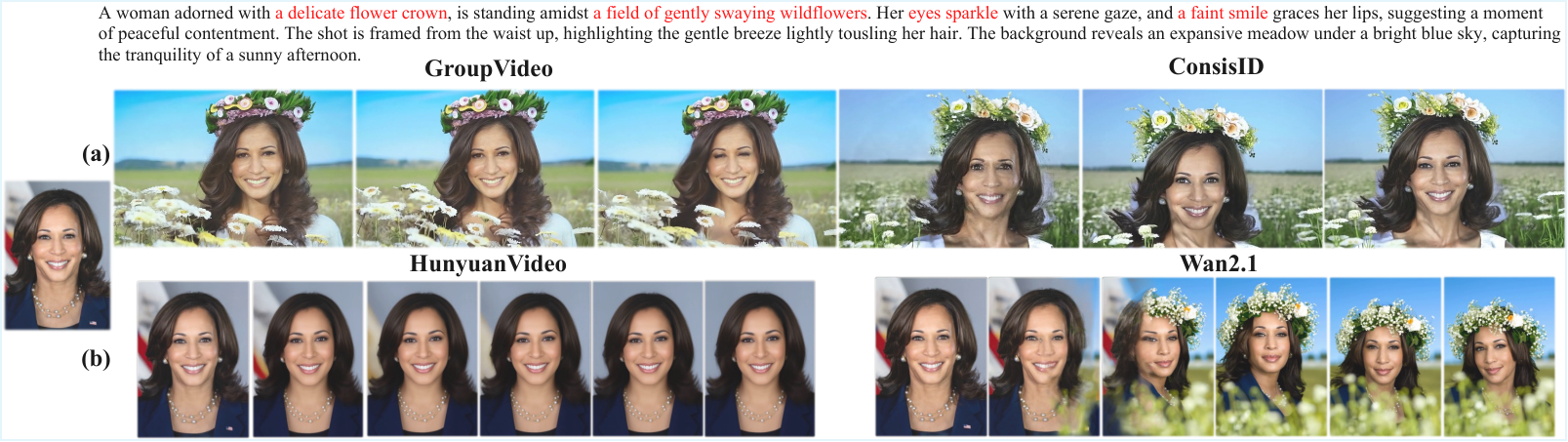}
   \caption{\textbf{Single-ID customization comparison} with the state-of-the-art methods ConsisID~\cite{yuan2024identity}, HunyuanVideo~\cite{kong2024hunyuanvideo} and Wan2.1~\cite{wan}.}
   \label{fig:6}
\end{figure*}

\begin{figure*}[t]
  \centering
   \includegraphics[width=1.0\linewidth]{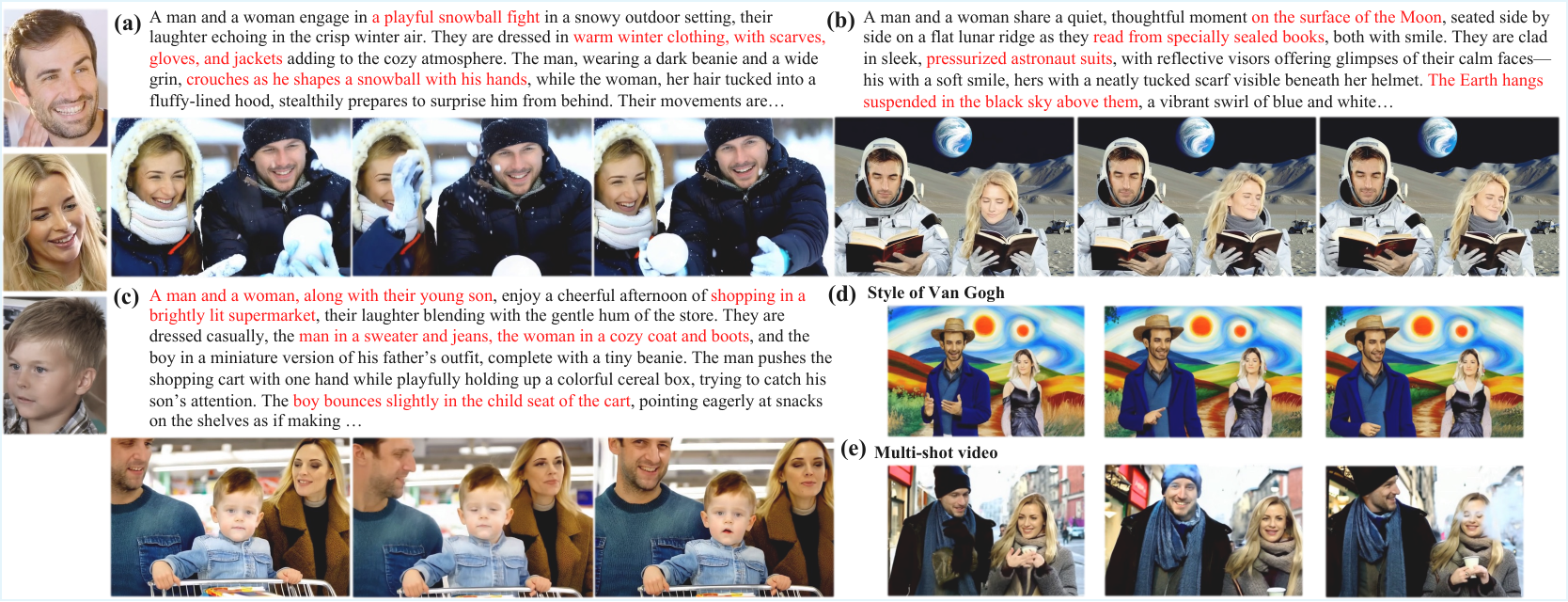}
   \caption{\textbf{More visual results} with (a) dynamic motion, (b) atypical backgrounds, (c) more individuals, (d) stylization, (e) multi-shots.}
   \label{fig:7}
\end{figure*}

\subsection{Experimental Settings}
\textbf{Implementation Details.}
We utilize CogVideoX-5B~\cite{yang2024cogvideox} as the foundational model for our framework. During training, each video is processed into continuous sequences of 49 frames, with each frame resized to a resolution of 480 $\times$ 720. The model was pre-trained on 400,000 internal samples and then fine-tuned on our 20,000 high-quality videos for task-specific performance. The total number of training steps is set to 160k for the first stage and 80k for the second stage, with a batch size of 2. Both stages employ the AdamW optimizer with a learning rate of 1e-5. The probability $\alpha$ for applying the bounding box constraint and the weighting factor $\lambda$ for the mask regularization are set to 0.5 and 0.1, respectively.
During training, alongside our multi-person dataset, we also incorporate single-person videos with a certain probability to enhance the model's generalization capability. 

\textbf{Evaluation.} 
\textcolor{black}{
We evaluate our method against state-of-the-art ID-preserved video generation models, including ID-Animator~\cite{he2024id}, ConsisID~\cite{yuan2024identity}, and multi-ID customization methods, Ingredients~\cite{fei2025ingredients}, Concat-ID~\cite{zhong2025concat} and MAGREF~\cite{deng2025magref}. 
Specifically, ID-Animator is implemented on top of AnimateDiff; ConsisID, Ingredients, and Concat-ID are built upon CogVideoX; and MAGREF is developed based on Wan-2.1.
}
Due to the lack of a dedicated multi-human evaluation dataset, we curated a test set comprising 50 facial images and 40 diverse prompts that equally represent various races, genders and ages, covering a wide range of expressions, actions and backgrounds for comprehensive assessment. The evaluation is conducted on a total of 500 test cases, with face images extracted from online videos and prompts designed with GPT-4.

\subsection{Main Results}

\textbf{Quantitative Results.}
Tab.~\ref{table:2} summarizes the quantitative results, where evaluation metrics are categorized into two dimensions: visual quality and conditional consistency. To measure identity preservation, we compute FaceSim~\cite{schroff2015facenet}, the pairwise face similarity between generated frames and real face images by calculating feature similarity in specific regions. For text alignment, CLIPScore~\cite{hessel2021clipscore} is utilized to evaluate the coherence to textual instructions. For video quality, we employ Fréchet Inception Distance (FID)~\cite{heusel2017gans} to assess the overall visual fidelity. Furthermore, Aesthetic Quality and Dynamic Degree from VBench~\cite{huang2024vbench} are adopted to measure visual appeal and motion dynamics, respectively.

\textcolor{black}{
As demonstrated in Tab.~\ref{table:2}, compared with methods based on AnimateDiff and CogVideoX (the first four rows), GroupVideo exhibits superior performance in text alignment, video quality, and dynamic representation, with Dynamic Degree improved by 12.0\% and FID improved by 8.25\%, indicating our enhanced video fidelity and naturalness. Even compared with MAGREF built on Wan-2.1, GroupVideo achieves better performance on all metrics except dynamic degree.} 
Although Ingredients and ConsisID achieve higher scores in face similarity due to their ``copy-paste" phenomena, our approach maintains consistent identity preservation throughout the entire video duration, accompanied by natural facial expression.
\textcolor{black}{
In the single-ID setting, although GroupVideo is mainly trained on multi-person data, it still achieves visual quality and condition alignment comparable to ConsisID, which is specifically designed for single-identity personalized video generation.
}

\textbf{Qualitative Results.}
Beyond the examples shown in Fig.~\ref{fig:1}, we present multi-identity comparative results in Fig.~\ref{fig:5}. For single-ID customization methods, we concatenate facial images as the input to provide multi-human references. As shown in the figure, ID-Animator struggles to maintain distinct identities, often blending multiple identity features, and exhibits almost no noticeable motion in the generated scenes. For DiT-based models, both ConsisID and Ingredients suffer from pronounced ``copy-paste" effect, with characters showing stiff expressions and unnatural movements. Furthermore, Ingredients produces outputs with noticeable distortions, such as red artifacts on the girl's hand in case 2 and a blackened arm in case 3. 
\textcolor{black}{
Concat-ID exhibits lower identity consistency and visual stability than the other methods. For example, in case 2, the girl’s skin tone and identity differ substantially from the input; in case 3, the results show obvious distortions and temporal jitter.
Although the Wan-2.1 backbone improves the motion dynamics of MAGREF, its text alignment is noticeably weaker. For instance, the “white V-neck t-shirt” in case 1 and “take a series of selfies” in case 3 are not reflected in the generated videos. Moreover, MAGREF tends to preserve ID-irrelevant attributes from the reference images (e.g., the man’s clothing in cases 1 and 3), which significantly compromises generation accuracy.
}
In contrast, the proposed GroupVideo consistently generates realistic and high-fidelity videos, achieving robust preservation of multiple identities and natural motion dynamics.

\begin{figure*}[t]
  \centering
  \includegraphics[width=1.0\textwidth]{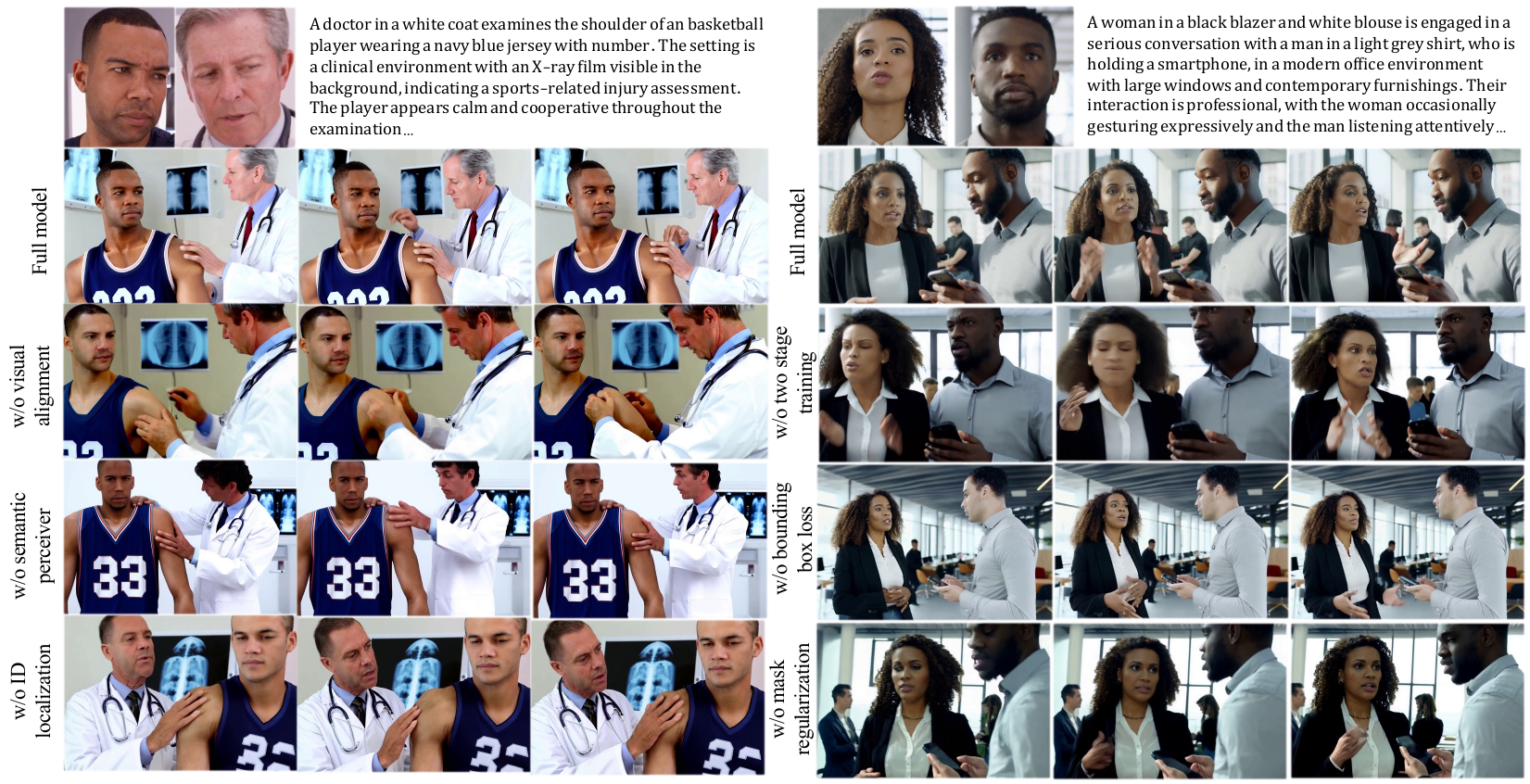}
   \caption{\textbf{Examples for ablation studies.} \textit{Left}: Ablation on structural modules including visual alignment, semantic perceiver and ID localization module. \textit{Right}: Ablation on loss functions and training strategies. Zoom in for more details.}
   \label{fig:8}
\end{figure*}

\begin{table}
\setlength{\tabcolsep}{11pt}
    \centering
    \caption{\textbf{User study results.} The table presents the average user ratings across four dimensions.}
    \label{table:3}
{
    \begin{tabular}{@{}lcccc@{}}
    \toprule
    Method   & \makecell{Identity\\Similarity}   & \makecell{Text\\Alignment}  & \makecell{Visual\\Quality} & \makecell{Motion\\Degree} \\
    \midrule
    ID-Animator   & 2.43 & 5.67 & 4.07 & 3.25 \\
    ConsisID & 6.46 & \underline{8.29} & \underline{6.72} & 5.85 \\
    Ingredients  & \textbf{6.75}  & 7.61  & 6.20 & \underline{6.33} \\
    \textbf{GroupVideo}   &  \underline{6.54}  &  \textbf{8.64}  & \textbf{6.93} & \textbf{6.79} \\ 
    \bottomrule
    \end{tabular}
}
\end{table}

\textbf{Single-ID comparison \& More visual results.} GroupVideo is not limited to multi-ID customization, while also maintaining good consistency in single-character scenarios. Here, we compare our approach with the state-of-the-art single-ID customization method ConsisID~\cite{yuan2024identity} as well as two leading video generation models, HunyuanVideo~\cite{kong2024hunyuanvideo} and Wan2.1~\cite{wan}. As shown in Fig.~\ref{fig:6}, GroupVideo naturally integrates the target character with the background, whereas Consis-ID exhibits noticeable artifacts around the character’s edges. Regarding the two image-to-video generation methods, both HunyuanVideo and Wan2.1 preserve the input image as the first frame. Besides, HunyuanVideo fails to generate video results that follow the given instructions, while Wan2.1 achieves a transition towards the target prompt but produces a character whose identity changes significantly. The quantitative comparison of single-ID customization with ConsisID is listed in Tab.~\ref{table:2}. In order to further demonstrate the generalization capabilities of GroupVideo,  diverse samples are presented in Fig.~\ref{fig:7}, including our results with dynamic motion, atypical background, three individuals, stylization and multi-shots video. The aforementioned results in diverse scenarios demonstrate the generalization capability of GroupVideo for ID-preserving customization tasks, effectively balancing semantic understanding and visual generation.

\textbf{User Study.}
To complement our quantitative metrics, we conduct a user study to evaluate the perceptual quality of the generated results. Each participant is asked to assess 100 video clips in terms of identity preservation, text alignment, video quality, and motion dynamics. Participants should rate each aspect on a scale of 1 to 10, with results summarized in Tab.~\ref{table:3}. A total of 74 valid responses are collected. As shown by the overall preference scores, GroupVideo achieves the highest ratings across nearly all evaluation dimensions, demonstrating its superior perceptual quality in human assessments. Although Ingredients achieves a higher user score in FaceSim due to its ``copy-paste" mechanism, it compromises the realism and authenticity.

\begin{table}
\setlength{\tabcolsep}{5pt}
    \centering
    \caption{\textbf{Ablation study.} FaceSim, CLIPScore, and FID are calculated to assess identity consistency, text alignment and video quality, respectively.}
    \label{table:4}
    {
    \begin{tabular}{@{}lccc@{}}
    \toprule
    Method   & FaceSim\cite{schroff2015facenet}$\uparrow$  & CLIPScore~\cite{hessel2021clipscore}$\uparrow$ & FID~\cite{heusel2017gans}$\downarrow$ \\
    \midrule
    w/o visual alignment    & 0.498       & 30.70  &  157.30 \\
    w/o semantic perceiver    & 0.531   & 28.55   &  168.87     \\
    w/o localization module   & 0.577    &  30.58  &   \textbf{131.06}   \\
    w/o two stage training    &  0.662    & 29.86   &  172.13  \\ 
    w/o bounding box loss   &  0.648    & 31.02   &  147.09  \\ 
    w/o mask regularization   &  \underline{0.710}  & \textbf{32.66}  &  141.64   \\ 
    \textbf{GroupVideo}  & \textbf{0.733} & \underline{31.95} & \underline{136.94}   \\
    \bottomrule
    \end{tabular}
    }
\end{table}

\subsection{Ablation Study}
\textbf{Key Modules.}

As illustrated in the left portion of Fig.~\ref{fig:8}, we conduct an ablation study to analyze the impact of key components of our GroupVideo: visual identity alignment module, semantic perceiver and localization module. (1) As shown in the second row, the absence of visual identity alignment, which means removing the face latent input of VAE, leads to a significant degradation in identity fidelity. This verifies that visual identity alignment is necessary for preserving identity information. (2) In the third-row example, the player’s face exhibits an obvious “copy-and-paste” artifact and the motion appears stiff; moreover, the doctor fails to maintain consistent identity. It indicates that the semantic perceiver is proven to be vital for generating reasonable actions and avoiding the "copy-paste" phenomenon. (3) The role of the ID localization module is demonstrated at the bottom of the figure, which displays a clear identity confusion (the doctor’s face is replaced by the player’s). The model tends to produce distinct identity blending artifacts without the ID location module.

\begin{figure*}[ht!]
  \centering
  \includegraphics[width=0.95\textwidth]{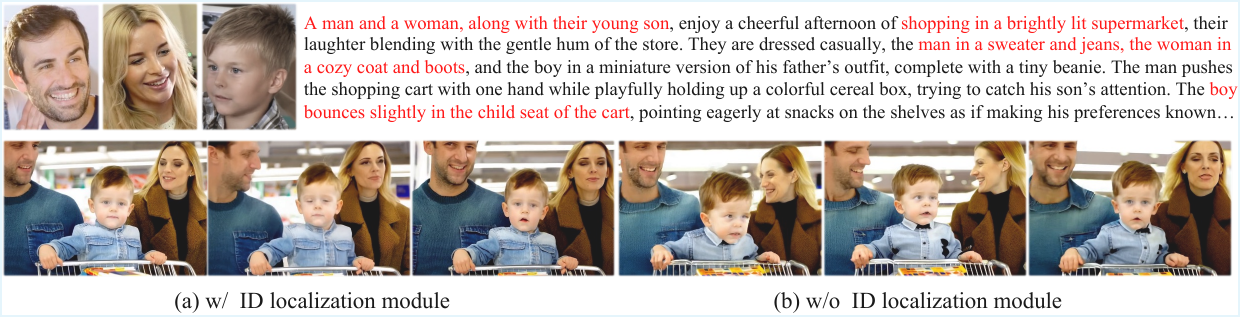}
  \caption{\textcolor{black}{Ablation study of the ID localization module with three individuals.}}
  \label{fig:9}
\end{figure*}

\begin{table}[t]
\centering
\textcolor{black}{
\setlength{\tabcolsep}{4.5pt}
    \caption{\textcolor{black}{Comparison of computational resources for ID customized video generation.}}
    \label{table:5}
 {
    \begin{tabular}{@{}lcccc@{}}
    \toprule
    Method   & Base Model & \makecell{Training \\Speed(s/step)}  & \makecell{Inference \\Time(s)} & \makecell{Inference \\Memory(GB)} \\
    \midrule
    ConsisID        & CogVideoX-5B &  24.6 & 531 & 9.7 \\
    Ingredients     & CogVideoX-5B &  -  & 552 & 31.0 \\
    Concat-ID       & CogVideoX-5B &  27.7  &  535 & 17.8  \\
    \textbf{GroupVideo} & CogVideoX-5B & 26.2 & 550 & 12.5 \\ 
    MAGREF          & Wan-2.1-14B      &  -  & 1400 & 72.0 \\
    \bottomrule
    \end{tabular}
}}
\end{table}

To further validate the robustness of the ID localization module, we conduct an ablation study in a three-character scenario. 
As shown in Fig.~\ref{fig:9}, model without the ID localization module produces slightly distorted and blurred faces compared to the variant with ID localization, leading to degraded overall identity consistency. Moreover, the overall visual quality is mildly compromised. These observations indicate that the ID localization module remains effective when handling more individuals, without a noticeable performance drop.

These observations are further supported by the quantitative metrics presented in the top three rows in Tab.~\ref{table:4}, which validate the importance of each component. Although the introduction of the ID localization module results in a slight trade-off in video quality, it significantly enhances face similarity. Collectively, these modules ensure the coherent injection of multiple identities and enhance the alignment between multimodal embeddings, improving the overall quality and consistency of the generated videos.

\begin{table}[t]
\centering
\textcolor{black}{
\setlength{\tabcolsep}{2.5pt}
    \caption{\textcolor{black}{Analysis of the computational overhead introduced by GroupVideo.}}
    \label{table:6}
 {
    \begin{tabular}{@{}lcccc@{}}
    \toprule
Training Modules & Stages & \makecell{Training \\Parameters(GB)} & \makecell{Training \\Speed(s/step)}  & \makecell{Training \\Memory(GB)} \\
    \midrule
baseline & \multirow{5}{*}{1st}  &  12.43  & 24.6 & 38.6 \\
+ multi-ID projection   &     &  12.43  & 24.8 & 39.5 \\
+ semantic perciever  &  &  12.70  & 25.3  &  40.3 \\
+ ID localization &  &  12.93  & 26.1  &  43.5 \\
+ bounding box loss   &  &  12.93  & 26.2 &  43.7 \\
    \midrule
only ID localization &  \multirow{2}{*}{2nd} & 0.41 &  21.1  &  11.1 \\
+ mask regularization        &    &   0.41  & 21.3 &  11.4 \\
    \bottomrule
    \end{tabular}
}}
\end{table}

\textbf{Training strategy and loss function.}
The right portion of Fig.~\ref{fig:8} highlights the impact of training strategies and loss functions. (1) As shown in the second row, single-stage training without second-stage finetuning results in unstable optimization and noticeable artifacts, whereas our proposed progressive two-stage training effectively regulates the optimization focus, thereby enhancing the temporal consistency of the model. (2) Compared with our full model, the results in the third and fourth rows exhibit varying degrees of degraded identity consistency. This indicates that both the bounding-box loss in Stage I and the mask regularization in Stage II contribute to more robust identity preservation. Notably, the results without the bounding-box loss sacrifice more identity details, indicating that the effect of bounding box constraints is more visually pronounced, as it provides explicit spatial guidance for identity localization. 

As illustrated in Tab.~\ref{table:4}, while the result without mask regularization achieves superior CLIPScore (32.66 vs. 31.95), this apparent advantage in text alignment stems from reduced spatial constraints - the unregularized model freely adapts non-facial regions to better match textual prompts, often at the cost of identity distortion. 
\textcolor{black}{
As shown in the last row, our full method strategically tolerates minor CLIPScore degradation to achieve significant gains in FaceSim (+2.3\%) and FID (+4.7), prioritizing identity preservation over literal prompt adherence. 
}
This trade-off proves essential for ID-consistent generation, where facial fidelity outweighs peripheral text alignment.

\subsection{\textcolor{black}{Computational Analysis}}

Our model is trained on 8 NVIDIA H20 GPUs (80 GB). With a batch size of 1, the minimum GPU memory usage is about 40 GB during the pre-training stage and only around 10 GB during the second-stage fine-tuning. As a result, the two-stage training increases the overall training time by about only 20\%, and our method does not impose high hardware requirements.

In Tab.~\ref{table:5}, we provide a detailed comparison of training speed, inference speed, and memory consumption across several ID-personalized video generation methods.
\textcolor{black}{Under the same hardware setting, GroupVideo requires 26.2 s/step during the main training stage, which is close to ConsisID (24.6 s/step) and Concat-ID (27.7 s/step). During inference, GroupVideo takes 550 s per sample with 12.5 GB memory, remaining comparable to other CogVideoX-based methods and substantially more efficient than Wan-2.1-based MAGREF (1400 s, 72.0 GB).
The results show that GroupVideo achieves training and inference speeds comparable to methods based on the same backbone, while requiring substantially less inference memory than Ingredients and Concat-ID, and offering a clear improvement over Wan-2.1-based methods.}

\textcolor{black}{
We further analyze the computational overhead introduced by GroupVideo. Compared with the baseline model ConsisID, the additional cost mainly comes from the newly introduced components, including multi-identity feature projection, the semantic perceiver, and the ID localization module, as well as two additional losses: the bounding box loss and the mask loss. Importantly, the second stage is lightweight, since only the localization module is optimized while the backbone DiT and the alignment modules are frozen. Therefore, the proposed progressive optimization does not lead to a proportional increase in training cost.}

\textcolor{black}{
Detailed computational overhead statistics for the above components, including the number of trainable parameters, training speed, and training memory consumption, are presented in Tab.~\ref{table:6}. In the leftmost column, each row adds the corresponding component based on the configuration in the previous row. All experiments report the average results over 100 training steps with batch size set to 1 on 8 H20 GPUs. As shown, the newly introduced modules, such as the multi-ID projection, semantic perciever, and ID localization module, increase the number of trainable parameters to some extent, but do not impose substantial additional computational overhead.
In addition, the bounding box loss in the 1st stage and the mask regularization loss in the 2nd stage introduce negligible parameter overhead, as they only act as auxiliary supervision terms rather than additional generative branches. 
These results indicate that the proposed framework achieves improved multi-identity controllability with moderate and practical computational overhead.
}

\section{Conclusion}
\label{sec:5}
In this paper, we propose GroupVideo, a unified zero-shot framework for multi-identity customized video generation. 
A novel framework is introduced to integrate multiple identity features into video generation through multi-modal identity alignment and an ID localization module.
Leveraging our curated multi-person video dataset and a progressive two-stage training strategy, GroupVideo achieves consistent and efficient optimization of multi-identity preservation. 
Extensive experiments demonstrate that GroupVideo achieves superior multi-identity fidelity with realistic dynamics, setting a new benchmark for identity-aware video synthesis.

\vfill

\end{document}